Title Page

# aSAGA: Automatic Sleep Analysis with Gray Areas

Matias Rusanen M.Sc.[1,2]*, Gabriel Jouan Ph.D.[5], Riku Huttunen M.Sc.[1,2], Sami Nikkonen Ph.D.[1,2], Sigríður Sigurðardóttir B.Sc.[5], Juha Töyräs Ph.D.[1,3,4], Brett Duce Ph.D.[7], Sami Myllymaa Ph.D.[1,2], Erna Sif Arnardottir Ph.D.[5,6], Timo Leppänen Ph.D.[1,2,4], Anna Sigridur Islind Ph.D.[8], Samu Kainulainen Ph.D.[1,2], Henri Korkalainen Ph.D.[1,2]

[1]Department of Technical Physics, University of Eastern Finland, Kuopio, Finland
[2]Diagnostic Imaging Center, Kuopio University Hospital, Kuopio, Finland
[3]Science Service Center, Kuopio University Hospital, Kuopio, Finland
[4]School of Electrical Engineering and Computer Science, The University of Queensland, Brisbane, Australia
[5]Reykjavik University Sleep Institute, School of Technology, Reykjavik University, Reykjavik, Iceland
[6]Landspitali–The National University Hospital of Iceland, Reykjavik, Iceland
[7]Sleep Disorders Centre, Princess Alexandra Hospital, Brisbane, Australia
[8]Department of Computer Science, Reykjavik University, Reykjavik, Iceland

**Corresponding author**

Matias Rusanen, M.Sc., Department of Technical Physics, University of Eastern Finland,

Yliopistonranta 1, P.O. BOX 1627 (Canthia), FI-70211 Kuopio, Finland.

E-mail: matias.rusanen@uef.fi

**Conflict of interests**

ESA reports honoraria from ResMed, Nox Medical, Jazz Pharmaceuticals, Linde Healthcare, Alcoa – Fjardaral, Wink Sleep, Apnimed and Vistor. She is also a member of the Philips Medical Advisory Board. Other authors declare no conflict of interest relevant to this study.

**Author contributorship**

Conceptualization: MR, GJ, RH, ESA, ASI, TL, SM, and SK; Methodology: MR, GJ, RH, HK, and SK; Analysis: MR and GJ; Writing – Original Draft: MR, GJ, SN, SS ; Writing – Review and Editing: All authors.

## Abstract

State-of-the-art automatic sleep staging methods have already demonstrated comparable reliability and superior time efficiency to manual sleep staging. However, fully automatic black-box solutions are difficult to adapt into clinical workflow and the interaction between explainable automatic methods and the work of sleep technologists remains underexplored and inadequately conceptualized. Thus, we propose a human-in-the-loop concept for sleep analysis, presenting an automatic sleep staging model (aSAGA), that performs effectively with both clinical polysomnographic recordings and home sleep studies. To validate the model, extensive testing was conducted, employing a preclinical validation approach with three retrospective datasets; open-access, clinical, and research-driven. Furthermore, we validate the utilization of uncertainty mapping to identify ambiguous regions, conceptualized as gray areas, in automatic sleep analysis that warrants manual re-evaluation. The results demonstrate that the automatic sleep analysis achieved a comparable level of agreement with manual analysis across different sleep recording types. Moreover, validation of the gray area concept revealed its potential to enhance sleep staging accuracy and identify areas in the recordings where sleep technologists struggle to reach a consensus. In conclusion, this study introduces and validates a concept from explainable artificial intelligence into sleep medicine and provides the basis for integrating human-in-the-loop automatic sleep staging into clinical workflows, aiming to reduce black-box criticism and the burden associated with manual sleep staging.

**Key terms:**

Sleep Staging, Human-in-the-Loop, Explainable Artificial Intelligence, Convolutional Neural Networks

# 1. Introduction

Manual sleep staging constitutes a fundamental analysis conducted in sleep clinics worldwide [1]. This process involves the examination of specific patterns in electroencephalography (EEG), electrooculography (EOG), and chin electromyography (chin EMG) within 30-second segments i.e., epochs of sleep recordings. Each epoch is subsequently categorized into one of five sleep stages: Wake, non-rapid eye movement sleep stages 1-3 (N1, N2, or N3), or rapid eye movement (REM) sleep. The established scoring rules are universally applied to guide this task [2]. Despite the diligent efforts invested in manual sleep staging, the presence of ambiguous regions persists, characterized by instances in which expert scorers are unable to reach a consensus and unintentionally introduce errors [3]–[5].

Uncertain areas with discrepancies between scorers, i.e. gray areas, of sleep staging exist for multiple reasons. Firstly, the scoring rules employed for sleep staging have faced criticism for insufficient scientific foundation and the potential for interpretive variations [6]–[8]. Additionally, scoring practices differ between sleep centers, as each institution usually adopts its own set of national or internal practices in addition to the traditionally used American Academy of Sleep Medicine scoring manual [9], [10]. Moreover, the simplification of scoring sleep stages into 30-second epochs can lead to confusions, particularly during transitions between stages [11]. Furthermore, sleep recordings are susceptible to artifacts, which can impact the accuracy of sleep stage analysis. In some instances, these artifacts may mimic characteristic patterns observed in EEG, EOG, or EMG traces specific to sleep stages [12]. While skilled sleep technologists can perform sleep staging relatively accurately, even under such challenging circumstances [13], the manual sleep staging necessitates substantial effort and an extensive allocation of work hours [14].

To alleviate the challenges associated with manual sleep staging, numerous automatic sleep staging solutions have been devised [15], [16]. These solutions have already been integrated into diverse polysomnographic (PSG) analysis software and are sometimes employed for preliminary scoring, which may speed up and increase the reliability of the manual sleep staging process [17]. Machine learning (ML) -based methods, representing the cutting-edge in this domain, demonstrate comparable performance to expert scorers in sleep staging, particularly in specific datasets where the models were trained on [15], [16], [18]–[24]. However, based on current clinical recommendations, automatic sleep analysis tools cannot be solely relied upon, and their results necessitate thorough manual review [25]. Only a few solutions have been introduced for highlighting the uncertain regions of automatic sleep analysis [19], [21], [26]–[28], and the concept warrants further validation. Especially, further studies are needed to test the generalizability of the methods across different datasets and recordings conducted on various medical sensor setups. Moreover, conceptualization and discussion regarding the interaction between automatic and manual scoring remains limited [3], [25]. This discussion would be crucial to consider for the human-in-the-loop approaches in sleep analysis where humans play a crucial role in guiding and validating the ML system's decisions, ensuring that they are clinically reasonable and aligned with human values [29].

This paper presents a novel conceptualization of the interaction between manual and automatic sleep staging (Figure 1). The primary objective of this approach is to enhance sleep staging accuracy while simultaneously reducing the workload associated with manual sleep staging. We hypothesize that this can be accomplished by automatically identifying and highlighting the most uncertain regions of automatic sleep staging, commonly referred to as gray areas. To realize this concept, we propose a model called aSAGA (Automatic Sleep Analysis with Gray Areas). This model integrates a state-of-the-art automatic sleep staging method that exhibits generalizability across different recording types [30], along with explainability through uncertainty mapping. The uncertainty mapping is based on the hypnodensities [24], [31], representing the probabilities of being in each sleep stage for each epoch. The aim is to assess whether the gray areas identified by the model correspond to regions where discrepancies arise between automatic and manual scoring as well as between sleep technologists. Furthermore, we aim to compare the effectiveness of hypnodensity-based uncertainty metrics introduced in the literature and discuss the possible strategies to define the gray areas.

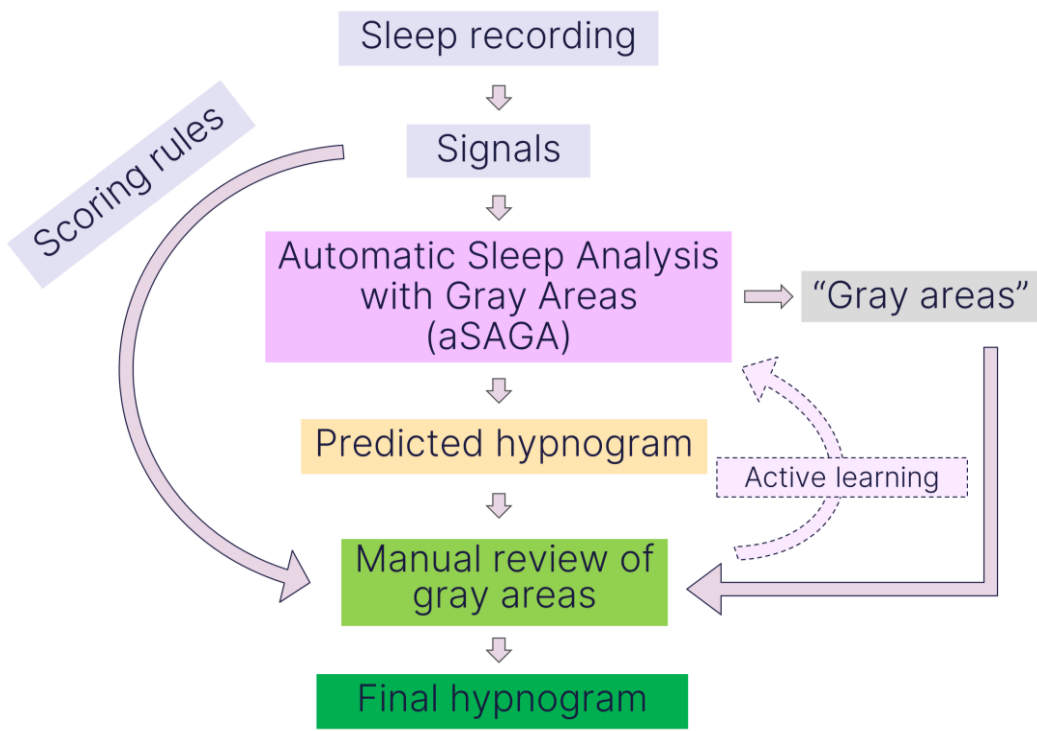

**Figure 1:** *The proposed human-in-the-loop sleep staging concept for the interaction of manual and automatic methods. In this approach, the aSAGA model gives a prediction of the hypnogram and highlights uncertain regions, the gray areas, that need manual re-evaluation. In this paper, we aimed to evaluate the aSAGA model with retrospective data. In future work, active learning can be incorporated as a feedback loop to the proposed approach.*

## 2. Methods

### *2.1 Datasets*

#### *2.1.1 Clinical PSG recordings*

The clinical in-laboratory PSG recordings used in the training of the automatic sleep staging model are described in detail in a previous work [20]. These recordings form a heterogeneous clinical dataset of suspected obstructive sleep apnea (OSA) patients collected at the Princess Alexandra Hospital (Brisbane, Australia) during the years 2015–2017 (Supplementary Table S1-S3). These recordings were conducted at the hospital's sleep laboratory with Compumedic (Abbotsford, Australia) Grael 4K PSG system with practices in line with the AASM guidelines [2]. The Institutional Human Research Ethics Committee of the Princess Alexandra Hospital reviewed and approved a research proposal for the use of this data (HREC/16/QPAH/021 and LNR/2019/QMS/54313). We excluded recordings with less than one hour of total sleep time and divided the data into training ($n$=710), validation ($n$=78), and test set ($n$=88). Demographic information of the data is presented in the supplementary material.

#### *2.1.2 Dreem Open Data*

The Dreem Open Data (DOD) is a documented and freely available open-access PSG dataset [32]. The data consists of 25 healthy participants (DOD-H) and 56 participants with OSA (DOD-O). The DOD-H data is collected at the French Armed Forces Biomedical Research Institute's (IRBA) Fatigue and Vigilance Unit using Siesta PSG devices (Compumedics, Victoria, Australia). The data included 13 EEG derivations (C3-M2, F4-M1, F3-F4, F3-M2, F4-M2, F4-O2, F3-O1, FP1-F3, FP1-M2, FP1-O1, FP2-F4, FP2-M1, and FP2-O2) as well as derivations for right and left EOG. The DOD-O data was collected at the Stanford Sleep Medicine Center using a Somno HD PSG device (Somnomedics). This data included 8 EEG derivations (C3-M2, C4-M1, F3-F4, F3-M2, F4-O2, F3-O1, FP1-M2, and FP1-F3) along with the left and right EOG. In both datasets, each record was scored by five expert sleep technologists (>5 years of clinical/scoring experience) from three different sleep centers. Further details on DOD data can be found in the original article [32]. An end-to-end example of how to use this data for training an automatic sleep staging model is available at https://github.com/UEF-SmartSleepLab/sleeplab-format/tree/main/examples/dod_sleep_staging.

#### *2.1.3 Self-applied PSG Data*

The recordings were collected at Akershus University Hospital in Norway and the Reykjavik University Sleep Institute in Iceland as a part of the Sleep Revolution Horizon 2020 project (ethical permissions 21-070 (Iceland) and 2017/2161 (Norway)) [33], [34]. The goal was to get a balanced cohort of healthy subjects and different levels of sleep apnea severity. A total of 50 participants received the self-applied PSG (Self-Applied Somnography (SAS), Nox Medical, Reykjavik, Iceland) equipment to take home for one night, along with written and video instructions on the setup. The SAS data comprised four EEG (AF4-E3E4, AF3-E3E4, AF7-E3E4, and AF8-E3E4) derivations and four EOG (E1-E4, E2-E3, E2-AFZ, and E3-AFZ) derivations that were used in the analysis. Demographic information of the data is presented in the supplementary material (Table S4).

Ten expert sleep technologists from seven different sleep centers scored each of the 50 recordings. Each scorer used Noxturnal Research version 6.1.0 (Nox Medical, Reykjavik, Iceland) software to first run an integrated automatic analysis and then inspected and scored/corrected the scoring following the AASM 2.6 scoring manual [35]. This scoring protocol was used to correspond to common clinical practice. In the end, we obtained ten scorings for each recording. A majority score for sleep stages was generated by selecting the most-scored sleep stage for each epoch for the 10 scorers [13]. A tiebreaker method was used to choose the sleep stage with higher priority in the following order; Wake, N1, N2, N3, and REM. Data included 3.5% of tied epochs. As a specialty, the scorers were able to mark their scoring as uncertain, if they were unsure about the sleep stage but still labeled with the most likely sleep stage e.g. uncertain N1. Following this, 3936 epochs (10% of analyzed data) were scored with uncertainty by at least one of the ten scorers.

## *2.2 Automatic Sleep Staging*

The presented aSAGA model is a fully convolutional neural network that has shown state-of-the-art performance and good generalizability between different datasets in previous studies [20], [30]. The architecture of the model was inspired by previous works by Perslev *et al.* [18], [36] and by Ronneberger *et al*. [37]. The model architecture is described in detail in [20]. The codes for setting up the model are available at https://github.com/rikuhuttunen/psg-simultscoring-models.

In the aSAGA, we used a single-channel model which was first trained on EEG (C4-M1) and then finetuned with an EOG (E1-M2) channel using the clinical PSG recordings. This was done to have generalizability between EEG and EOG channels and to increase the compliance of the model for frontal EEG and EOG channels of self-applicable sensor setups. In the training, we utilized a recently published unified data format, called the *sleeplab format* [38], to have better reproducibility. The training of the model was conducted over 100 epochs. During each training epoch, each batch input to the model was created by sampling a one-hour segment of recording from 8 different subjects. The start times of the segments were randomly sampled from each recording. Model weights were optimized using the AdamW algorithm [39]. The categorical cross entropy was used as the loss function. Furthermore, instead of a constant learning rate, we used a learning rate scheduler [40]. The learning rate division factor was 2.0 after every 20 epochs and the peak learning rate was set to 0.01. After initial training, a constant learning rate of 0.001 was used in the finetuning. Otherwise, the finetuning was performed similarly to the initial training. The best model was returned after training according to validation accuracy. Training parameters and model parameters used in the training and finetuning are available in detail at https://github.com/matias-olavi/aSAGA.

In the aSAGA, all EEG and EOG signals are pre-processed similarly to training data; bandpass filtered (0.3–35 Hz), re-sampled to 64 Hz, and normalized with an interquartile range. Standard preprocessing functions of the *sleeplab format* extractor were used (available in GitHub). Then all signals were run through the single-channel model to create an ensemble prediction of the hypnodensities i.e., the probability of each sleep stage for each 30-second epoch. Following this, the hypnodensities were averaged over the channels and the sleep stage with the highest probability was included in the final hypnogram.

*2.3 Uncertainty Mapping – the Gray Areas*

Gray areas were derived from manual and automatic sleep staging similarly using uncertainty mapping. In the aSAGA, averaged hypnodensities from the ensemble predictions were used as the probability of the sleep stage in the studied epoch. In manual sleep staging from ten scorers, the probability for each sleep stage can be defined as how many of the ten scorers scored that sleep stage. Let $p_i$, where i = 1, 2,…,5 correspond to each of the five sleep stages, denote the probabilities or the hypnodensities for each 30-second epoch. We can estimate the uncertainty of the model from the hypnodensities with different metrics described in the literature [19], [26], [29]. In the present paper, we compared the following five metrics:

1. Least confidence sampling

$$U_L = 1 - max(p_I) * \frac{n}{n-1},$$

   where $I = \{all\ sleep\ stages\}$ and $n$ is the number of sleep stages.

2. The margin of confidence sampling (between the two most confident labels)

$$U_M = 1 - (max(p_I) - max(p_J)),$$

   where $I = \{all\ sleep\ stages\}$, and $J = \{all\ sleep\ stages\ except\ the\ most\ probable\}$.

3. The ratio of confidence (between the two most confident labels)

$$U_R = max(p_I)/max(p_J),$$

   where $I = \{all\ sleep\ stages\}$, and $J = \{all\ sleep\ stages\ except\ the\ most\ probable\}$.

4. Coefficient of unlikeability

$$U_U = 1 - \sum_{i=1}^{n} p_i^2$$

5. Classification entropy

$$U_E = -\frac{\sum_{i=1}^{n} p_i \log_2 p_i}{\log_2 n},$$

where were assumed $0 * \log_2(0) = 0$.

## *2.4 Validation analysis*

To validate the single-channel automatic sleep staging model, we first used the test set of clinical PSG recordings collected at the Princess Alexandra Hospital to show the sleep staging performance from the single EOG and EEG channels. Typical performance metrics such as accuracy, Cohen kappa, and F1-scores were computed in an epoch-wise manner to measure the overall sleep staging performance. Sleep stage-specific classification performance was evaluated by computing precision, recall, and F1-score for each class separately. Furthermore, we visualized the scorings between the automatic and manual sleep staging with a confusion matrix.

To validate the generalizability of the aSAGA, we evaluated the sleep staging performance with the Dreem Open Data. The performance of the automatic method was tested with healthy subjects and patients with OSA separately against the consensus of five scorers. In this validation, we utilized all EEG and EOG channels available in the dataset to create the ensemble prediction described in Section 2.3. Finally, the automatic sleep staging performance of the aSAGA was tested with the self-applied PSGs against the consensus of the ten scorers. No finetuning of the model was performed *i.e.*, the performance was evaluated in a direct transfer manner with all datasets (sections 2.1.2 and 2.1.3).

Following the sleep staging performance validation, we tested the gray area concept of the aSAGA with two different analyses. First, we tested if excluding the most uncertain epochs of the prediction increased the automatic sleep staging performance when evaluated against the consensus manual scoring. This was done to evaluate if the gray areas corresponded to the areas of the recording where the confusion between manual and automatic sleep staging occurs. This analysis was performed using the uncertainty metrics proposed in Section 2.4 and varying the percentage of epochs considered as gray areas from 1% to 95% of the whole dataset based on the rank order of the uncertainty values. This validation was done using the Dreem Open Data, and the validation protocol is available at https://github.com/matias-olavi/aSAGA.

Second, we tested whether the gray areas of the aSAGA capture those areas of the self-applied PSGs where the ten scorers had the most confusion between each other. The testing was performed using the coefficient of unlikeability, which is suitable for capturing the variability of the manual scorings. A threshold of 0.6 was used for the uncertainty metric to separate gray areas from non-gray areas with both, the 10 scorers and model hypnodensity series. The data for this analysis was split into two parts, the epochs that were marked as uncertain by any of the 10 scorers, and the remaining data (not marked as uncertain by any of the scorers). The data consisting of epochs marked as uncertain represent gray areas that are known to exist in the sleep stage, whereas gray areas of data consisting of manually scored epochs that were not marked as uncertain represent more the unknown uncertainties. Finally, the binary classification performance between the gray and non-gray areas derived from manual and automatic sleep staging was computed. Confusions in the gray area classification between these two methods were visualized with a confusion matrix.

## 3. Results

### *3.1 Sleep Staging Performance*

#### *3.1.1 Clinical PSG data – test set (n=88)*

The single-channel model performed well against manual sleep staging and with similar overall metrics (accuracy = 81% and kappa = 0.74) for both EEG and EOG inputs (Table 1). Sleep stage-specific metrics showed that REM sleep was better detected from the EOG input than from the EEG input, with F1-scores of 0.88 and 0.83, respectively. In contrast, Wake was detected with similar consistency from EEG compared to EOG as an input, with F1-scores of 0.90 and 0.89, respectively. Most of the confusions between automatic and manual sleep staging occurred in the classification of non-REM sleep stages of N1, N2, and N3 (Figure 2).

**Table 1:** Performance metrics of the automatic sleep staging model in a test set of the clinical PSG data when single-channel inputs were used.

| Recording type | Subjects | Signal | Sleep Stage | Performance metrics | | | Number of epochs |
|---|---|---|---|---|---|---|---|
| | | | | Precision | Recall | F1-score | |
| Type I PSG | Suspected OSA (*n*=88) | EOG (E1-M2) | Wake | 0.91 | 0.87 | 0.89 | 26881 |
| | | | N1 | 0.51 | 0.33 | 0.40 | 6708 |
| | | | N2 | 0.75 | 0.84 | 0.80 | 25370 |
| | | | N3 | 0.80 | 0.79 | 0.80 | 9695 |
| | | | REM | 0.85 | 0.91 | 0.88 | 9567 |
| | | | Macro average | 0.76 | 0.75 | 0.75 | 78221 |
| | | | κ | | 0.74 | | 78221 |
| | | | Accuracy | | 81% | | 78221 |
| | | EEG (C4-M1) | Wake | 0.93 | 0.88 | 0.90 | 26881 |
| | | | N1 | 0.50 | 0.35 | 0.41 | 6708 |
| | | | N2 | 0.72 | 0.89 | 0.80 | 25370 |
| | | | N3 | 0.85 | 0.74 | 0.79 | 9695 |
| | | | REM | 0.91 | 0.77 | 0.83 | 9567 |
| | | | Macro average | 0.78 | 0.73 | 0.75 | 78221 |
| | | | κ | | 0.74 | | 78221 |
| | | | Accuracy | | 81% | | 78221 |

Metrics were computed against the manual sleep staging. The macro average is the unweighted average over the sleep stage-specific metrics. Abbreviations: epoch = 30-second signal segment, κ = Cohen's kappa, EEG = electroencephalography, EOG = electrooculography, OSA = obstructive sleep apnea, PSG = polysomnography, REM = rapid eye movement.

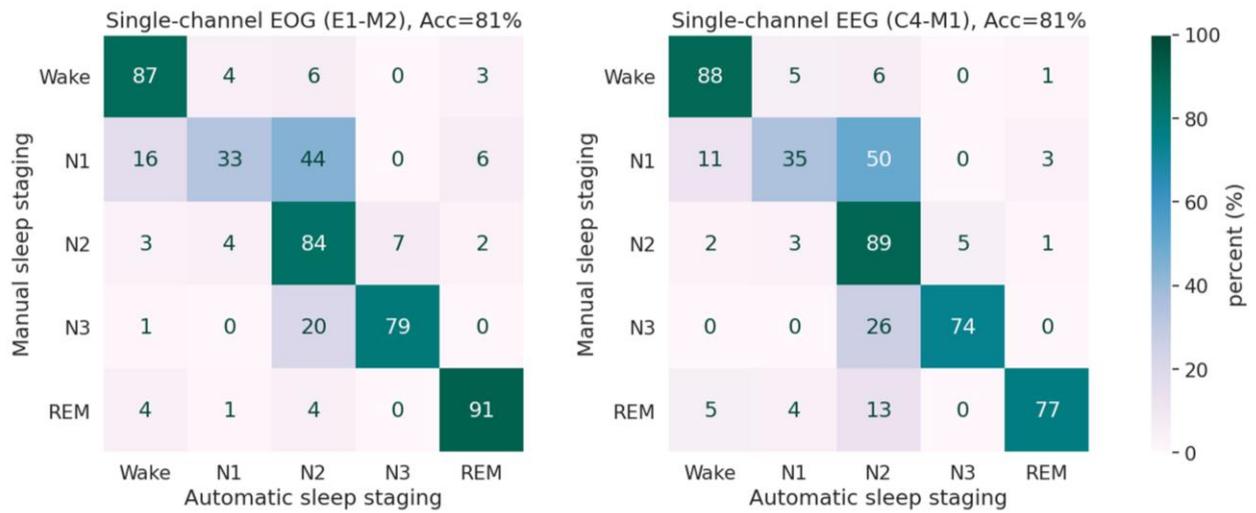

**Figure 2:** *Confusion matrixes between manual sleep staging and the single-channel outputs of automatic sleep staging with the aSAGA model. Performance was computed on the test set of the clinical polysomnographic recordings of suspected OSA patients (n=88). Abbreviations: Acc = Accuracy, aSAGA = Automatic Sleep Analysis with Gray Area, EEG = electroencephalography, EOG = electro-oculography.*

### 3.1.2 Dreem Open Data (n=25+56)

The aSAGA model showed high generalizability of the sleep staging performance to separate open-access datasets consisting of recordings of healthy subjects (DOD-H, *n*=25) and subjects with OSA (DOD-O, *n*=56). With DOD-H data, the aSAGA model reached 80% accuracy, and with DOD-O data the accuracy was 82% without any transfer learning approaches. Sleep stages Wake, N2, and REM were detected with a good agreement (F1-scores≥0.82) against the majority scoring (Table 2). Slightly lower performance was reached for the detection of N3 sleep, with F1-scores being 0.78 and 0.79 for DOD-H and DOD-O data, respectively. However, N1 sleep was detected with moderate agreement in both datasets.

**Table 2:** Performance metrics for the automatic sleep staging in Dreem Open Data when the aSAGA model was used.

| Recording type | Subjects | Signals | Sleep Stage | Performance metrics | | | Number of epochs |
|---|---|---|---|---|---|---|---|
| | | | | Precision | Recall | F1-score | |
| Type I PSG | Healthy (*n*=25) | All EEG + EOG channels | Wake | 0.87 | 0.77 | 0.82 | 3037 |
| | | | N1 | 0.71 | 0.29 | 0.41 | 1505 |
| | | | N2 | 0.83 | 0.83 | 0.83 | 11879 |
| | | | N3 | 0.65 | 0.97 | 0.78 | 3514 |
| | | | REM | 0.87 | 0.78 | 0.82 | 4727 |
| | | | Macro average | 0.79 | 0.73 | 0.73 | 24662 |
| | | | κ | | 0.71 | | 24662 |
| | | | Accuracy | | 80% | | 24662 |
| | OSA (*n*=56) | | Wake | 0.90 | 0.79 | 0.84 | 10660 |
| | | | N1 | 0.50 | 0.42 | 0.46 | 2898 |
| | | | N2 | 0.85 | 0.86 | 0.86 | 26650 |
| | | | N3 | 0.71 | 0.89 | 0.79 | 5683 |
| | | | REM | 0.85 | 0.82 | 0.83 | 8306 |
| | | | Macro average | 0.76 | 0.76 | 0.76 | 54197 |
| | | | κ | | 0.74 | | 54197 |
| | | | Accuracy | | 82% | | 54197 |

Metrics were computed against the majority scores of manual sleep staging. Macro average is the unweighted average over the sleep stage-specific metrics. Abbreviations: aSAGA = Automatic Sleep Analysis with Gray Areas, epoch = 30-second signal segment, κ = Cohen's kappa, EEG = electroencephalography, EOG = electrooculography, OSA = obstructive sleep apnea, PSG = polysomnography, REM = rapid eye movement.

### 3.1.3 Self-applied PSGs

Manual sleep staging of the self-applied PSGs showed high variation in the agreement of the ten individual scorings against the majority score with accuracy ranging from 69% to 91% between the ten scorers (Figure 3). On bar with these manual scorings against the majority score, the aSAGA reached the accuracy of 83% against the majority scoring. High F1-scores (≥0.86) were reached for detecting N2, N3, and REM sleep (Table 3, Figure 4). The F1-score for detecting Wake was lower, however. This was due to a precision of 0.63, although a high recall of 0.85 was reached. Similarly, as in the manual scorings, stage N1 was detected with only a fair agreement (F1-score=0.30), and mostly confused with Wake and N2 stages (Figure 4).

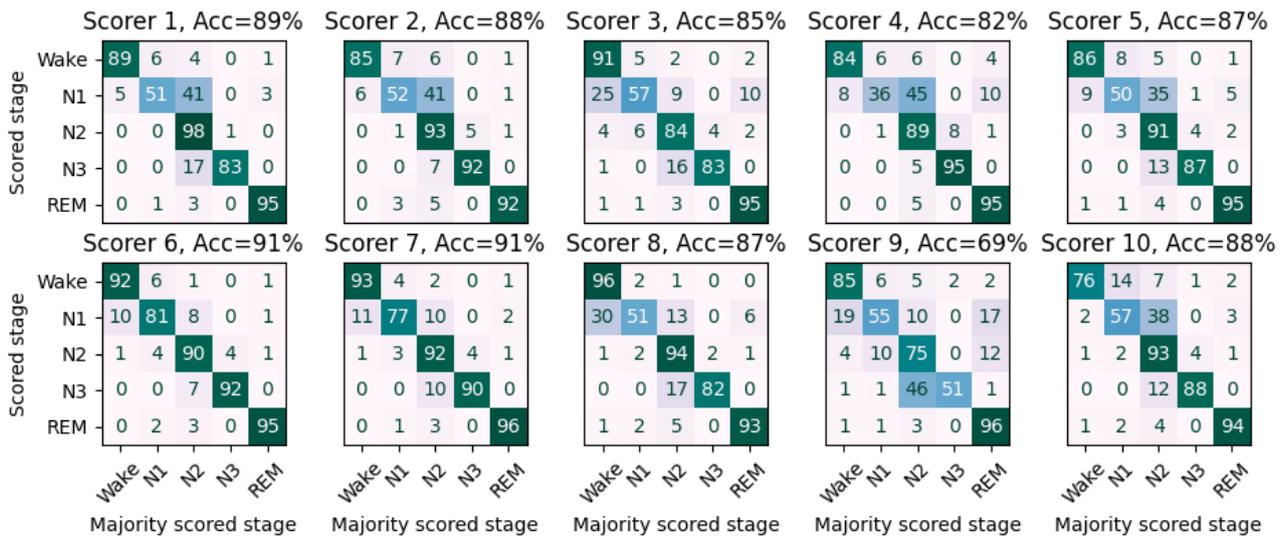

**Figure 3:** *Confusions between the ten individual scorings and the majority score derived from manual sleep staging in self-applied polysomnographic recordings (n=50). Epochs manually marked as uncertain were excluded from the analysis. Abbreviations: Acc = Accuracy, REM = rapid eye movement.*

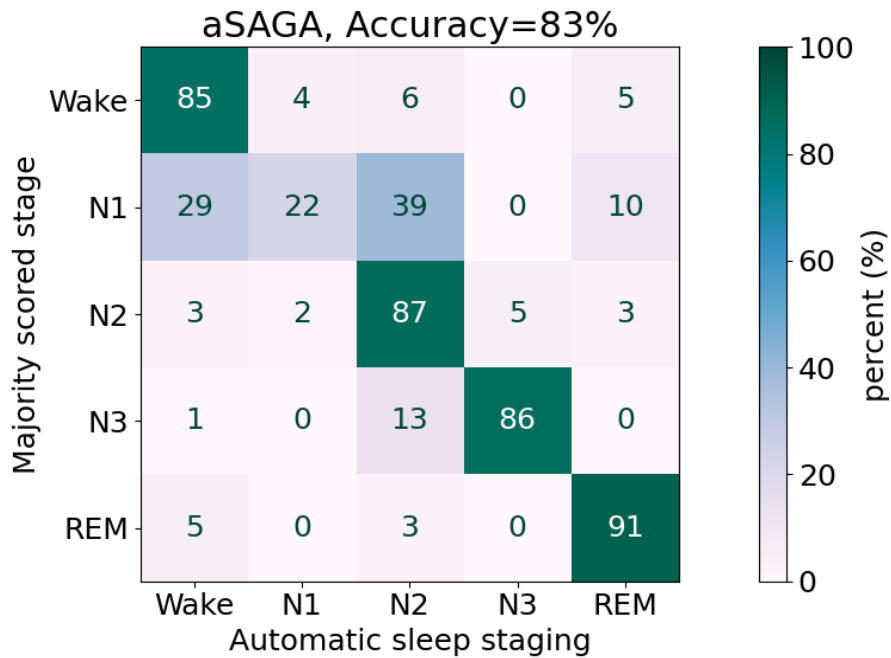

**Figure 4:** *Confusions between automatic sleep staging of aSAGA model and the majority score derived from manual sleep staging in self-applied home sleep recordings. Epochs manually marked as uncertain were excluded from the analysis. Abbreviations: aSAGA = Automatic Sleep Analysis with Gray Areas, REM = rapid eye movement.*

**Table 3:** Performance metrics for the automatic sleep staging in self-applied polysomnographic (PSG) recordings when the aSAGA model was used.

| Recording type | Subjects | Signals | Sleep Stage | Performance metrics | | | Number of epochs |
|---|---|---|---|---|---|---|---|
| | | | | Precision | Recall | F1-score | |
| Self-applied PSG | Healthy to severe OSA (*n*=50) | All EEG + EOG channels | Wake | 0.63 | 0.85 | 0.73 | 3183 |
| | | | N1 | 0.50 | 0.22 | 0.30 | 2303 |
| | | | N2 | 0.88 | 0.87 | 0.87 | 17443 |
| | | | N3 | 0.87 | 0.86 | 0.86 | 6136 |
| | | | REM | 0.85 | 0.91 | 0.88 | 6168 |
| | | | Macro average | 0.75 | 0.74 | 0.73 | 35233 |
| | | | κ | | 0.75 | | 35233 |
| | | | Accuracy | | 83% | | 35233 |

Metrics were computed against the majority scores of manual sleep staging. Epochs manually scored with uncertainty were excluded from the analysis. Macro average is the unweighted average over the sleep stage-specific metrics. Abbreviations: aSAGA = Automatic Sleep Analysis with Gray Areas, epoch = 30-second signal segment, κ = Cohen's kappa, EEG = electroencephalography, EOG = electrooculography, SAS = Self-Applied Somnography (Nox Medical).

### *3.2 Gray Area Validity*

#### *3.2.1 Increase in Sleep Staging Performance*

Automatic sleep staging performance increased effectively as a function of the amount of excluded data i.e., the gray areas, based on all of the compared uncertainty metrics in the DOD-O data (Figure 5). Differences in increased sleep staging performance were negligible between the different uncertainty metrics on less than 10% of excluded data (Figure 5). However, classification entropy ($U_E$) showed less increase in sleep staging performance between 10% and 75% of excluded data, than the other metrics. Similarly, the coefficient of unlikeability ($U_U$) increased the sleep staging performance slightly less between 10% and 30% of excluded data than the more direct measures of hypnodensities ($U_L$, $U_M$, and $U_R$).

Furthermore, confusions in sleep staging between manual and automatic scoring were captured effectively into gray areas (Figure 6). For example, when considering 40% of the data as gray areas, over 80% of the confusion between manual and automatic sleep staging was captured with $U_L$, $U_M$, $U_R$, and $U_U$ metrics. However, entropy-based metrics $U_U$ and $U_E$ showed slightly worse overall performance than the more direct measures of hypnodensities $U_L$, $U_M$, and $U_R$.

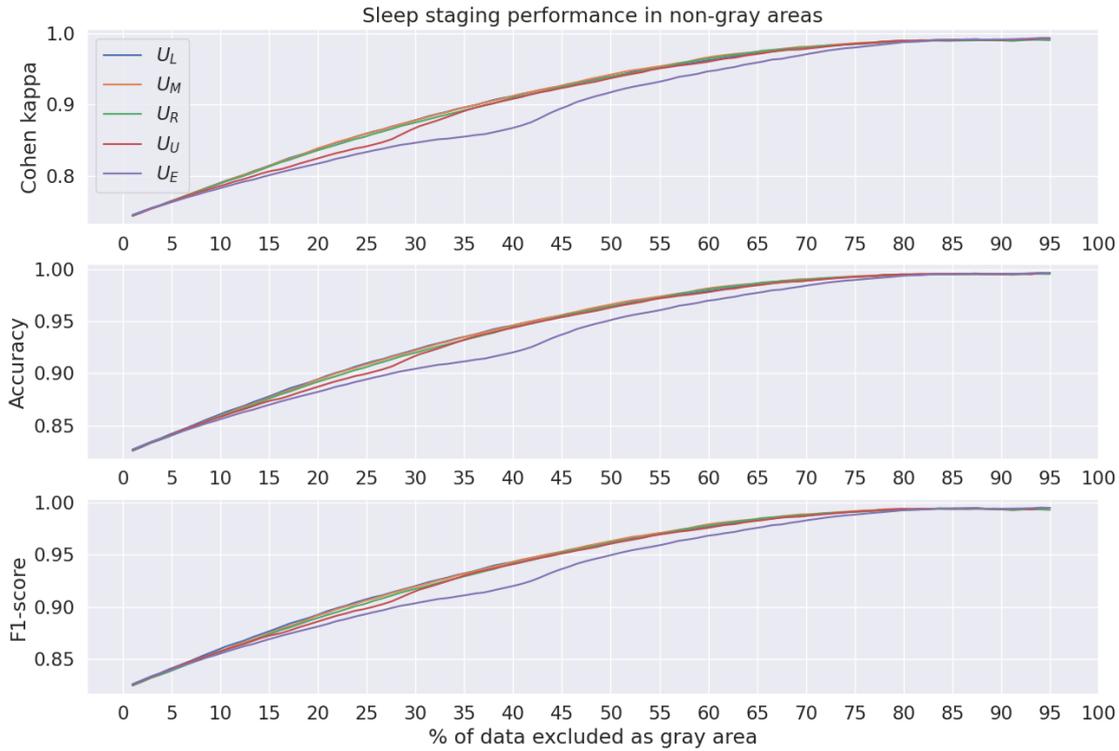

**Figure 5:** *Performance metrics of the aSAGA model in automatic sleep staging as a function of the amount of data excluded as gray areas. Gray areas were defined based on different hypnodensity-based uncertainty metrics (U) to compare the effectiveness of these measures. The compared metrics were the least confidence $U_L$, the margin of confidence $U_M$, the ratio of confidence $U_R$, the coefficient of unlikeability $U_U$, and the classification entropy $U_E$. F1-score is the weighted average of class-wise F1-scores. Dreem Open Data with obstructive sleep apnea subjects was used in the analysis (n=56 full-night polysomnographies). Abbreviations: aSAGA = Automatic Sleep Analysis with Gray Areas.*

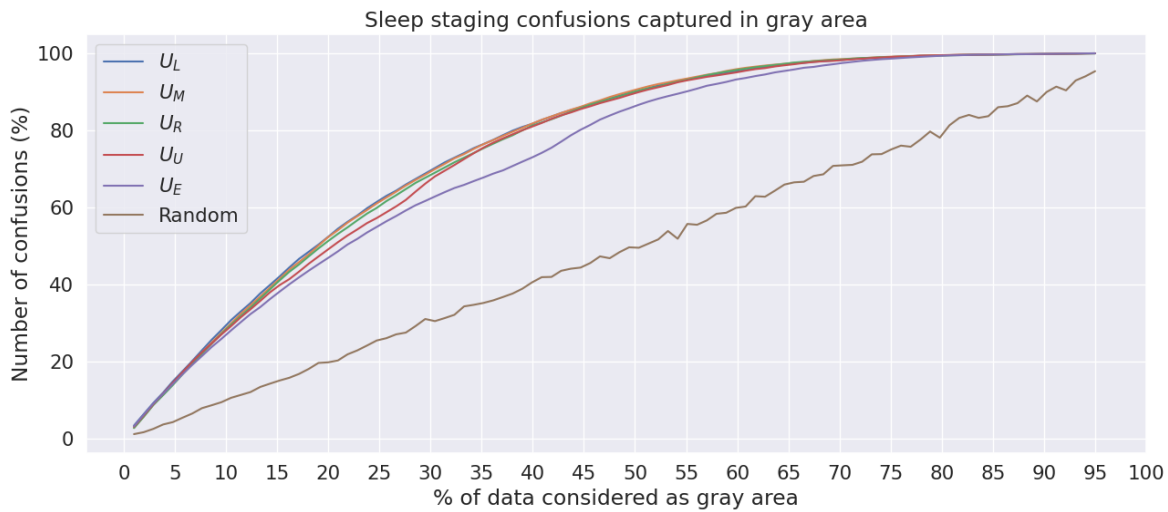

**Figure 6:** *Percentage of sleep staging confusion between automatic and manual majority scoring captured in the model gray areas. Gray areas were defined based on different hypnodensity-based uncertainty metrics (U) to compare the effectiveness of these measures. The compared metrics were the least confidence $U_L$, the margin of confidence $U_M$, the ratio of confidence $U_R$, the coefficient of unlikeability $U_U$, and the classification entropy $U_E$. Dreem Open Data with obstructive sleep apnea subjects was used in the analysis (n=56 full-night polysomnographies). Abbreviations: aSAGA = Automatic Sleep Analysis with Gray Areas*

*3.2.2 Agreement with Manual Scoring Confusions in Self-applied PSG*

An example of the hypnodensity series and gray areas derived from manual and automatic sleep staging of one recording is shown in Figure 7. Defining the gray areas similarly from the 10 scorers and the automatic model's hypnodensity (with threshold $U_U > 0.6$) series resulted in a median of 2.4% and 10.3% of analyzed epochs as gray areas in the 50 recordings, respectively.

When all epochs were split into known uncertainties and unknown uncertainties, the model gray areas agreed better with the manual scoring gray areas within the known uncertainties than within the unknown uncertainties (Figure 8). In both cases, the model captured most of the manual scoring gray areas i.e., 54% within unknown uncertainties and 61% within the known uncertainties.

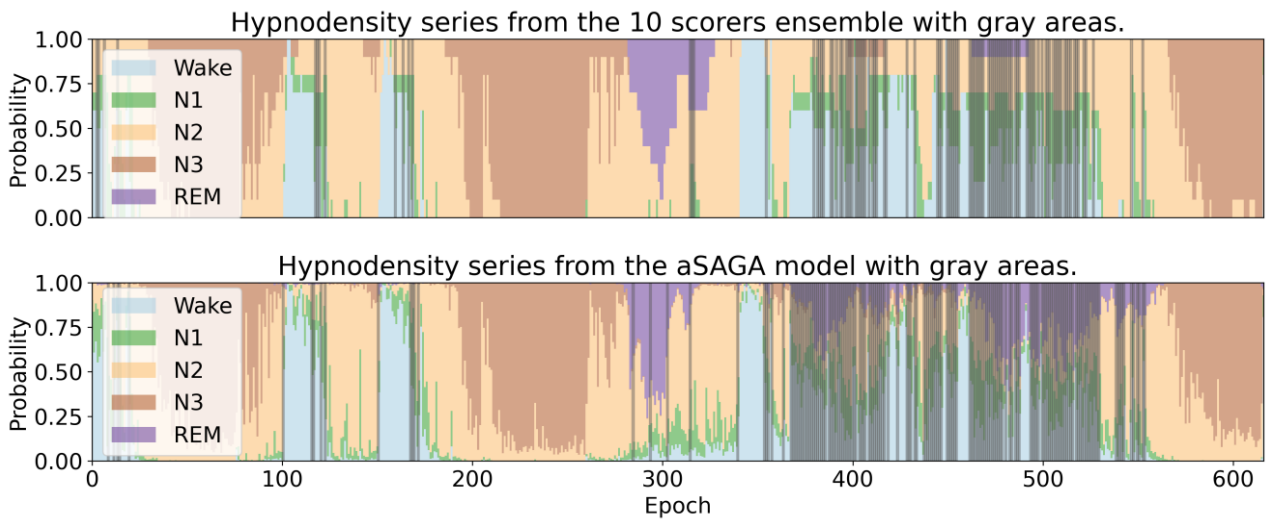

**Figure 7:** *An example of hypnodensity series and gray areas derived from manual sleep staging confusions and aSAGA model outputs for one of the self-applied home sleep recordings. Hypnodensities are shown as stacked bars with the respective colors indicated in the legend. Gray areas are shown as shaded vertical bars of a gray color. Abbreviations: aSAGA = Automatic Sleep Analysis with Gray Areas.*

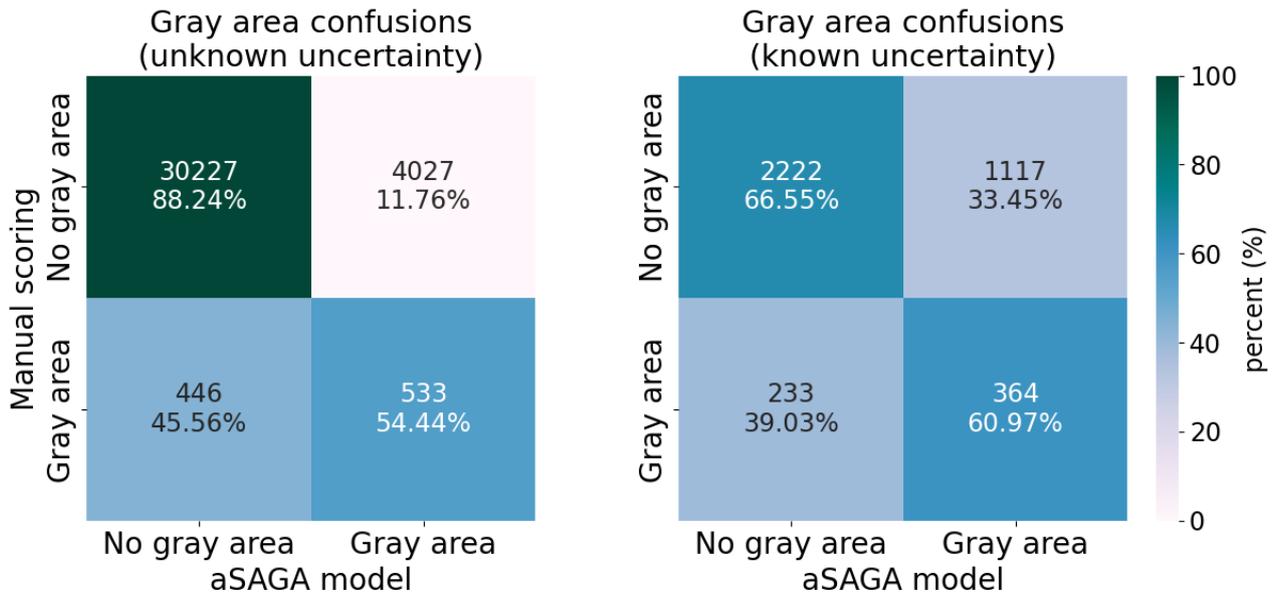

**Figure 8**: *Confusions in the gray areas derived from ten manual scorings and the aSAGA model output. Known uncertainties included epochs that were scored as uncertain at least by one of the ten scorers (10% of all data). Unknown uncertainties represent those areas of the recording that were scored with certainty by all 10 scorers. Abbreviations: aSAGA = Automatic Sleep Analysis with Gray Areas*

## 4. Discussion

This paper presents a novel human-in-the-loop concept for sleep staging that incorporates the integration of automatic analysis and manual review of the automatic output. The efficacy of various hypnodensity-based uncertainty metrics was evaluated in delineating the gray areas within automatic sleep staging and these gray areas compared with those observed in manual sleep staging. Through preclinical validation, we observed that the performance of the automatic sleep staging model was comparable to that of manual scorers in different recording types. Additionally, the gray areas identified from the automatic sleep staging captured most of the regions of the recording where significant discrepancies between multiple manual scorers and automatic sleep staging occurred.

Consequently, focusing on the manual revision of the gray areas alone could enhance the accuracy of sleep staging, while simultaneously minimizing the workload associated with fully manual sleep staging. By leveraging the concept of gray areas, the proposed approach offers a valuable user-assistive solution for improving the reliability of sleep staging, especially with unconventional sensor setups. At the same time, better explainability can reduce the skepticism related to the application of machine learning methods in clinical medicine [41], [42], which is one of the greatest challenges in implementing automatic PSG scoring at scale [43]. Nevertheless, further research is warranted to evaluate the clinical feasibility, incorporate active learning [29], and quantify the reduction in working hours achievable with this approach.

A few previous studies have investigated the uncertainty of automatic sleep staging, utilizing various approaches that should be discussed [19], [21], [26]–[28]. Phan *et al.* focused on a single entropy metric for hypnodensities and employed a fixed threshold to assess sleep staging accuracy in uncertain and certain epochs [19]. Kang et al. conducted a study involving the use of model uncertainty to enhance sleep staging accuracy, examining 20 recordings [26]. However, their findings revealed statistically significant improvement in Cohen's kappa only for REM vs. N3, rather than for agreement across all five stages. Fiorillo et al. and Hong et al. introduced sophisticated models capable of estimating the uncertainty of automatic sleep staging, aiming to capture discrepancies between manual and automatic approaches [21], [27]. Previous validation analyses have primarily focused on evaluating the extent of discrepancies between manual and automatic scoring within uncertain epochs, without considering the true gray areas associated with manual sleep staging. Only one study compared hypnodensities derived from multiple manual scorings against automatic scoring, revealing a strong correlation between the two [28]. However, none of these works have demonstrated the generalizability of their methods across different recording types. In this paper, we contribute to the field by considering both validation aspects and demonstrating the generalizability of our approach across diverse recording types. Additionally, as a novel contribution, we compare different uncertainty metrics and aim to stimulate discussion surrounding various strategies for defining the gray areas of automatic sleep analysis.

In this study, we defined gray areas using two different approaches. Firstly, when comparing uncertainty metrics, gray areas were defined as a percentage of the most uncertain epochs within the recording. Secondly, when comparing uncertainties between the automatic and multiple manual scorings in self-applied PSGs, gray areas were determined using a fixed threshold for the applied uncertainty metric.

Using the threshold method, the number of gray areas varied across recordings, reflecting the inherent variations in the difficulty of scoring different recordings e.g., healthy vs. OSA subjects [44]. It is natural for some recordings to present greater challenges in analysis than others, and theoretically, the threshold method should capture this. However, when employing a fixed value for uncertainty, the calibration of the model influences the number of gray areas. If the model is applied to data differing from the training data, higher uncertainties may be obtained, resulting in an increased number of gray areas. Now the model was trained on standard PSG signals and applied on the forehead EEG of the self-applied PSGs. This could explain why the threshold method yielded more gray areas based on model hypnodensities compared to ten scorer hypnodensities in self-applied PSGs (with a median percentage of 2.4% for manual scoring and 10.3% for the model).

Conversely, a fixed percentage of the most uncertain epochs as gray areas could be considered, and then the number of gray areas would remain constant for each recording. However, this approach also has its limitations. Recordings that pose greater challenges for the model's analysis would be treated the same as recordings that are easier to analyze. Consequently, some true gray areas might be overlooked in difficult-to-analyse recordings, or already correctly analyzed epochs might be highlighted in easier-to-analyse recordings, which would not be a good use of the manual scorer's time.

The most effective method to define the gray areas for the interaction between manual and automatic sleep staging remains uncertain. Ideally, the sensitivity for gray areas, such as the threshold for uncertainties or the percentage of most uncertain epochs, could be interactively adjusted for each recording, based on the expertise of the sleep technologist. This personalized approach could optimize the identification of gray areas and enhance the overall performance of the sleep staging process. Another approach would be to use, for example, automatic domain adaptation methods and to calibrate the model for each recording type separately [45]. Furthermore, model uncertainty is an active area of machine learning research, and novel ways to quantify it, such as Monte Carlo dropout, should be tested further in the field of sleep medicine [21], [29], [46].

In conclusion, this study sheds light on the concept of gray areas in the context of sleep staging and introduces a way to combine manual and automatic sleep staging. Through extensive evaluation and comparison, we have demonstrated that the proposed automatic sleep staging model, along with the identification of gray areas, can achieve similar performance to manual scorers. The two approaches employed for defining gray areas, namely the percentage-based method and the fixed threshold method, offer distinct advantages and trade-offs. However, the optimal approach for determining gray areas should be personalized, considering the expertise of sleep technologists and the specific characteristics of each recording. This paper provides novel tools and insights to effectively integrate explainable automatic sleep staging into clinical workflows, alleviating the burden of manual sleep staging and enhancing sleep staging accuracy.

## Data availability

The private clinical data from Princess Alexandra Hospital include sensitive medical information and can therefore not be made publicly available. The self-applied PSGs are stored in a safe medical data environment for research purposes, but not made publicly available. Dreem Open Data is publicly available. End-to-end examples on how to download the open access data and train a sleep staging model with this data are given at https://github.com/UEF-SmartSleepLab/sleeplab-format/tree/main/examples/dod_sleep_staging.

## Acknowledgments

Financial support for this study was provided by the European Union's Horizon 2020 research and innovation programme under grant agreement No 965417, by the Research Committee of the Kuopio University Hospital Catchment Area for the State Research Funding (Grants 5041807, 5041789, 5041794, 5041797, 5041803), by the Finnish Cultural Foundation through Kainuu Regional Fund, by the Emil Aaltonen Foundation, by the Research Foundation of the Pulmonary Diseases, by the Foundation of the Finnish Anti-Tuberculosis Association, by Tampere Tuberculosis Foundation, and by the NordForsk (NordSleep Project 90458) through Business Finland (Grant 5133/31/2018) and the Icelandic Research Fund.

The authors thank Sleep Technologists at the Charite-Universitaetsmedizin Berlin, Berlin, Germany; Instituti Clinici Scientific Maugeri Spa Societa Benefit, Pavia, Italy; Princess Alexandra Hospital, Brisbane, Australia; Reykjavik University Sleep Institute, Reykjavik, Iceland; Turku University Central Hospital, Turku, Finland; University of Gothenburg, Gothenburg, Sweden; and University of Lisbon, Lisbon, Portugal for manual scoring of the self-applied polysomnography recordings.

July, pp. 1–12, 2022, doi: 10.1093/sleep/zsac154.

[29]   R. Monarch, *Human-in-the-loop machine learning : active learning and annotation for human-centered AI*. Shelter Island: Manning.

[30]   M. Rusanen *et al.*, "Generalizable Deep Learning-based Sleep Staging Approach for Ambulatory Textile Electrode Headband Recordings," *IEEE J. Biomed. Heal. Informatics*, vol. 27, no. 8, pp. 1869–1880, 2023, doi: 10.1109/JBHI.2023.3240437.

[31]   P. Anderer, M. Ross, A. Cerny, R. Vasko, E. Shaw, and P. Fonseca, "Overview of the hypnodensity approach to scoring sleep for polysomnography and home sleep testing," *Front. Sleep*, vol. 2, 2023, doi: 10.3389/frsle.2023.1163477.

[32]   A. Guillot, F. Sauvet, E. H. During, and V. Thorey, "Dreem Open Datasets: Multi-Scored Sleep Datasets to Compare Human and Automated Sleep Staging," *IEEE Trans. Neural Syst. Rehabil. Eng.*, vol. 28, no. 9, pp. 1955–1965, Sep. 2020, doi: 10.1109/TNSRE.2020.3011181.

[33]   E. S. Arnardottir *et al.*, "The Sleep Revolution project: the concept and objectives," *J. Sleep Res.*, vol. 31, no. 4, pp. 1–8, 2022, doi: 10.1111/jsr.13630.

[34]   B. F. Sveinbjarnarson, L. Schmitz, E. S. Arnardottir, and A. S. Islind, "The Sleep Revolution Platform: a Dynamic Data Source Pipeline and Digital Platform Architecture for Complex Sleep Data," *Curr. Sleep Med. Reports*, vol. 9, no. 2, pp. 91–100, 2023, doi: 10.1007/s40675-023-00252-x.

[35]   R. B. Berry *et al.*, "AASM Manual for the Scoring of Sleep and Associated Events," Amer. Acad. Sleep Med., Darien, IL, USA, 2018. doi: 10.1016/j.carbon.2012.07.027.

[36]   M. Perslev, M. H. Jensen, S. Darkner, P. J. Jennum, and C. Igel, "U-Time: A fully convolutional network for time series segmentation applied to sleep staging," *arXiv*, pp. 1–19, 2019, [Online]. Available: https://arxiv.org/abs/1910.11162.

[37]   O. Ronneberger, P. Fischer, and T. Brox, "U-Net: Convolutional Networks for Biomedical Image Segmentation," in *Medical Image Computing and Computer-Assisted Intervention – MICCAI 2015*, 2015, pp. 234–241, doi: 10.1007/978-3-319-24574-4_28.

[38]   R. Huttunen, S. Kainulainen, S. Nikkonen, M. Rusanen, and H. Korkalainen, "UEF-SmartSleepLab/sleeplab-format." 2023, doi: 10.5281/zenodo.7861887.

[39]   I. Loshchilov and F. Hutter, "Decoupled weight decay regularization," 2019. [Online]. Available: https://arxiv.org/abs/1711.05101.

[40]   L. N. Smith, "A disciplined approach to neural network hyper-parameters: Part 1 -- learning rate, batch size, momentum, and weight decay," 2016. [Online]. Available: https://arxiv.org/abs/1803.09820.

[41]   P. Linardatos, V. Papastefanopoulos, and S. Kotsiantis, "Explainable ai: A review of machine learning interpretability methods," *Entropy*, vol. 23, no. 1, pp. 1–45, 2021, doi: 10.3390/e23010018.

[42]   C. Rudin, "Stop explaining black box machine learning models for high stakes decisions and use interpretable models instead," *Nat. Mach. Intell.*, vol. 1, no. 5, pp. 206–215, 2019, doi: 10.1038/s42256-019-0048-x.
22

# Supplementary

**Table S1**. Demographic information of test dataset subjects (*n*=88) in clinical PSG recordings data

|       | Sex | Age [years] | BMI [kg/m^2] | TST [minutes] | WASO [minutes] | N1% | N2% | N3% | REM% | AHI [1/h] |
|-------|-----|-------------|--------------|---------------|----------------|------|------|------|------|-----------|
| males | 51% |             |              |               |                |      |      |      |      |           |
| mean  |     | 55.62       | 34.57        | 291.70        | 120.74         | 13.71| 49.11| 19.48| 17.70| 24.35     |
| std   |     | 13.15       | 7.97         | 86.30         | 64.21          | 10.62| 11.93| 14.35| 7.29 | 28.16     |
| min   |     | 23          | 17.20        | 82.00         | 23.00          | 1.30 | 11.20| 0.00 | 0.00 | 0.20      |
| 25%   |     | 45          | 29.63        | 230.50        | 64.88          | 6.08 | 42.80| 8.20 | 13.33| 7.85      |
| 50%   |     | 56          | 33.65        | 298.75        | 116.50         | 12.35| 49.50| 18.85| 18.60| 15.00     |
| 75%   |     | 65          | 38.48        | 354.75        | 158.00         | 17.65| 57.30| 25.80| 21.53| 30.65     |
| max   |     | 85          | 66.20        | 480.50        | 326.50         | 68.40| 83.00| 72.00| 38.80| 144.80    |

BMI = Body Mass Index, N1-N3% = Percent of Stage N1-N3 sleep, TST = Total Sleep Time, REM% = Percent of Stage REM sleep, WASO = Wake After Sleep Onset, AHI = Apnea Hypopnea Index (events/hours of sleep).

**Table S2.** Demographic information of validation dataset subjects (*n*=78) in clinical PSG recordings data.

|       | Sex | Age [years] | BMI [kg/m^2] | TST [minutes] | WASO [minutes] | N1% | N2% | N3% | REM% | AHI [1/h] |
|-------|-----|-------------|--------------|---------------|----------------|------|------|------|------|-----------|
| males | 65% |             |              |               |                |      |      |      |      |           |
| mean  |     | 54.11       | 34.50        | 309.36        | 107.28         | 15.90| 50.24| 17.55| 16.50| 27.84     |
| std   |     | 14.74       | 8.78         | 84.96         | 57.66          | 12.40| 12.02| 11.99| 8.54 | 27.36     |
| min   |     | 20          | 21.70        | 89.00         | 13.00          | 1.50 | 20.70| 0.00 | 0.00 | 0.20      |
| 25%   |     | 43          | 27.95        | 253.00        | 61.88          | 7.70 | 43.13| 9.40 | 11.20| 6.05      |
| 50%   |     | 54          | 33.10        | 312.75        | 101.25         | 12.55| 50.35| 17.70| 16.60| 19.80     |
| 75%   |     | 65          | 38.68        | 362.13        | 150.38         | 22.15| 56.50| 24.30| 23.10| 44.40     |
| max   |     | 84          | 64.50        | 486.00        | 287.00         | 79.30| 89.20| 58.50| 38.50| 116.00    |

One subject did not want to report their sex and one subject had missing demographic data. BMI = Body Mass Index, N1-N3% = Percent of Stage N1-N3 sleep, TST = Total Sleep Time, REM% = Percent of Stage REM sleep, WASO = Wake After Sleep Onset, AHI = Apnea Hypopnea Index (events/hours of sleep).

**Table S3.** Demographic information of training dataset subjects (*n*=710) in clinical PSG recordings data.

|  | Sex | Age | BMI [kg/m^2] | TST [minutes] | WASO [minutes] | N1% | N2% | N3% | REM% | AHI [1/h] |
|---|---|---|---|---|---|---|---|---|---|---|
| males | 54% | | | | | | | | | |
| mean | | 54.21 | 35.84 | 303.06 | 112.12 | 14.93 | 48.16 | 19.79 | 17.12 | 24.13 |
| std | | 14.60 | 9.99 | 79.50 | 66.18 | 12.77 | 12.41 | 13.64 | 8.14 | 23.59 |
| min | | 17 | 0.00 | 67.50 | 7.50 | 0.00 | 5.50 | 0.00 | 0.00 | 0.00 |
| 25% | | 44 | 29.30 | 257.63 | 61.00 | 6.63 | 41.13 | 9.98 | 11.93 | 7.10 |
| 50% | | 55 | 34.70 | 309.00 | 101.25 | 10.80 | 47.90 | 18.50 | 17.15 | 15.60 |
| 75% | | 65 | 40.65 | 360.00 | 148.38 | 18.78 | 55.90 | 27.48 | 22.20 | 32.30 |
| max | | 88 | 76.20 | 531.50 | 407.00 | 87.10 | 87.20 | 89.30 | 47.10 | 143.10 |

BMI = Body Mass Index, N1-N3% = Percent of Stage N1-N3 sleep, TST = Total Sleep Time, REM% = Percent of Stage REM sleep, WASO = Wake After Sleep Onset, AHI = Apnea Hypopnea Index (events/hours of sleep).

**Table S4.** Demographic information of subjects (*n*=48) in self-applied PSG recordings data.

|  | Sex | Age | BMI [kg/m^2] | TST [minutes] | WASO [minutes] | N1% | N2% | N3% | REM% |
|---|---|---|---|---|---|---|---|---|---|
| males | 46% | | | | | | | | |
| mean | | 48.82 | 28.14 | 391.76 | 37.12 | 8.18 | 53.79 | 19.30 | 19.49 |
| std | | 17.21 | 6.26 | 79.37 | 37.50 | 7.14 | 8.97 | 9.69 | 7.36 |
| min | | 22 | 19.61 | 192.50 | 2.50 | 1.43 | 33.40 | 1.55 | 0.94 |
| 25% | | 32 | 29.30 | 347.00 | 13.75 | 4.20 | 48.46 | 12.23 | 15.82 |
| 50% | | 50 | 26.87 | 404.00 | 24.00 | 5.75 | 52.37 | 18.25 | 20.16 |
| 75% | | 61 | 31.45 | 444.12 | 45.62 | 9.99 | 60.38 | 25.20 | 23.58 |
| max | | 81 | 50.73 | 540.50 | 192.50 | 35.61 | 72.52 | 42.72 | 37.94 |

Two subjects had missing demographic data. Sleep stage percentages are computed from the majority sleep stage sequence. BMI = Body Mass Index, N1-N3% = Percent of Stage N1-N3 sleep, TST = Total Sleep Time, REM% = Percent of Stage REM sleep, WASO = Wake After Sleep Onset, AHI = Apnea Hypopnea Index (events/hours of sleep).